\documentclass[letterpaper, 10 pt, conference]{ieeeconf}  

\IEEEoverridecommandlockouts                              

\usepackage{epsfig}
\usepackage{graphicx}
\usepackage{amsmath}
\usepackage{amssymb}

\usepackage{multirow}
\usepackage{multicol}
\usepackage{mathtools}
\usepackage{graphicx}
\usepackage{caption}
\usepackage{array}
\usepackage{lipsum}

\usepackage[capitalize]{cleveref}
\crefname{section}{Sec.}{Secs.}
\Crefname{section}{Section}{Sections}
\Crefname{table}{Table}{Tables}
\crefname{table}{Tab.}{Tabs.}

\title{\LARGE \bf
DR-Pose: A Two-stage Deformation-and-Registration Pipeline for Category-level 6D Object Pose Estimation
}

\author{Lei Zhou$^{1*}$, Zhiyang Liu$^{1*}$, Runze Gan$^{1}$, Haozhe Wang$^{1,2\dagger}$ and Marcelo H. Ang Jr.$^{1}$
\thanks{*The authors contributed equally to this work.}
\thanks{$^{1}$Authors are with the Department of Mechanical Engineering, National University of Singapore, 117608, Singapore. {\tt\small \{leizhou, zhiyang, ganrunze, wang\_haozhe\}@u.nus.edu}, {\tt\small mpeang@nus.edu.sg}}%
\thanks{$^{2}$Haozhe Wang is with the Integrative Sciences and Engineering Programme, National University of Singapore Graduate School, 119077, Singapore.}%
\thanks{$^\dagger$Corresponding Author.}
}

\begin{document}

\maketitle
\thispagestyle{empty}
\pagestyle{empty}

\begin{abstract}
   Category-level object pose estimation involves estimating the 6D pose and the 3D metric size of objects from predetermined categories. While recent approaches take categorical shape prior information as reference to improve pose estimation accuracy, the single-stage network design and training manner lead to sub-optimal performance since there are two distinct tasks in the pipeline. In this paper, the advantage of two-stage pipeline over single-stage design is discussed. To this end, we propose a two-stage deformation-and-registration pipeline called DR-Pose, which consists of completion-aided deformation stage and scaled registration stage. The first stage uses a point cloud completion method to generate unseen parts of target object, guiding subsequent deformation on the shape prior. In the second stage, a novel registration network is designed to extract pose-sensitive features and predict the representation of object partial point cloud in canonical space based on the deformation results from the first stage. DR-Pose produces superior results to the state-of-the-art shape prior-based methods on both CAMERA25 and REAL275 benchmarks. Codes are available at https://github.com/Zray26/DR-Pose.git.

\end{abstract}
\section{Introduction}
\label{sec:intro}

The accurate estimation of object pose and size is crucial for a variety of real-world applications, including autonomous driving, augmented reality \cite{marchand_pose_2016}, scene understanding \cite{zhang_holistic_2021}\cite{nie_total3dunderstanding_2020}, and robotic manipulation \cite{deng_self-supervised_2020}. While most current object pose estimation networks focus on instance-level object pose estimation \cite{wang_densefusion_2019-1}\cite{he_ffb6d_2021-1}\cite{PVNET}\cite{xiang2018posecnn}, which requires exact instance CAD models and their sizes beforehand, this approach can be limiting in real-world scenarios where such information may not be available. To address this challenge, category-level object pose estimation is becoming increasingly popular as it enables the accurate estimation of the pose of unseen objects belonging to the same category.

\begin{figure}[t]
	\centering
	\includegraphics[width=\linewidth]{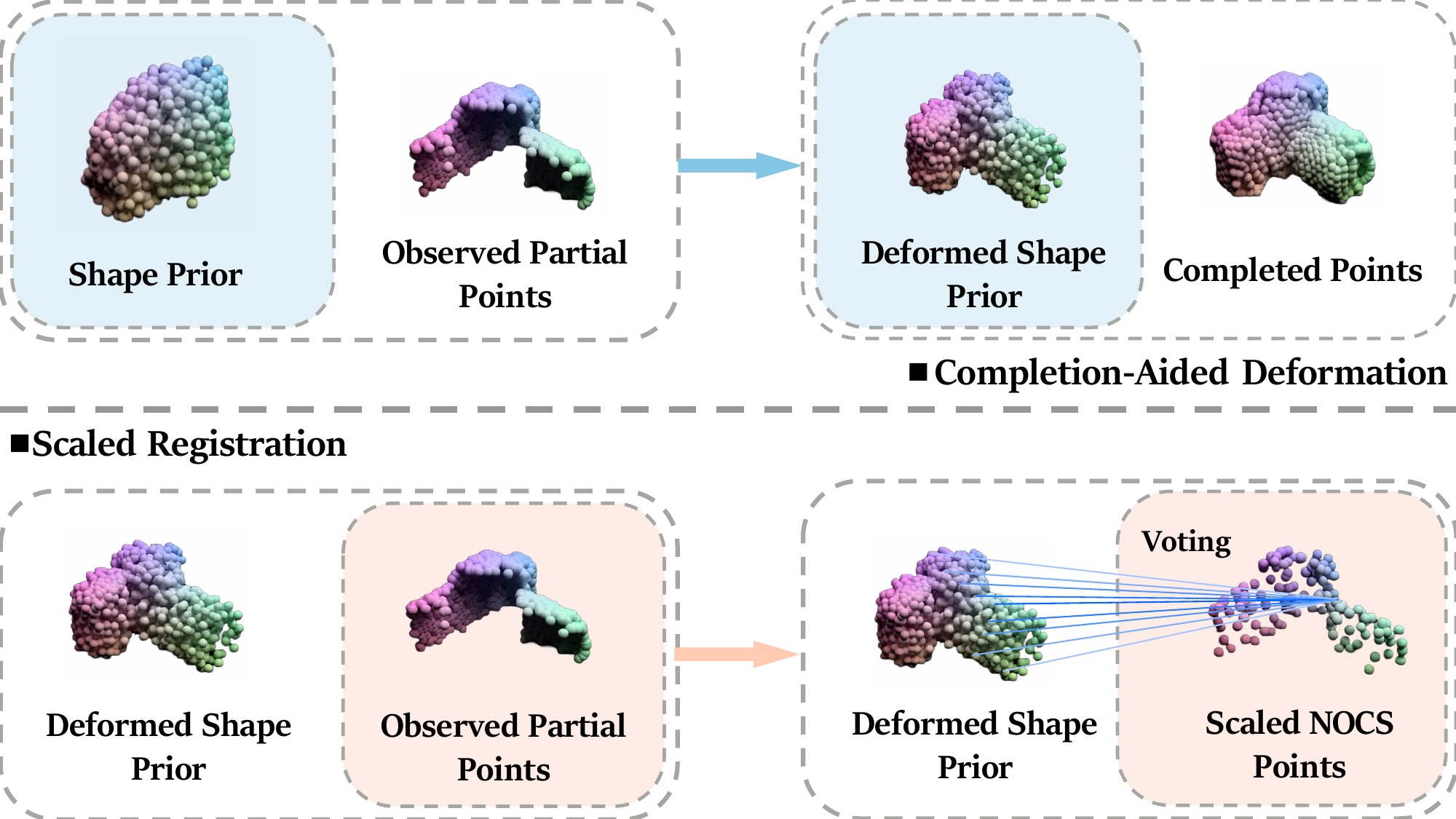}
	\caption{\textbf{DR-Pose} consists of two stages: completion-aided deformation and scaled registration. In the first stage, the partial point cloud of target object is completed, which is then taken as reference to deform the categorical shape prior to reconstruct the scale-invariant CAD model for the target object in the canonical space. In the second stage, taking the deformed shape prior as reference, coordinates of each observed object point are voted in the canonical space and further adjusted by the predicted scaling factors.}
	\label{teaser}	
\end{figure}

Recently, there has been a surge in learning-based approaches for category-level object pose estimation, which have demonstrated significant improvements over conventional method \cite{121791}. However, intra-class variations pose a significant challenge to these approaches. To handle intra-class variations, Normalized Object Coordinate Space (NOCS) \cite{NOCS} has been adopted to represent all instances of one category in a generalized implicit representation. Based on NOCS, SPD \cite{SPD} generates normalized categorical shape priors with auto-encoder to provide categorical prior information, which is utilized as reference to improve the accuracy of NOCS \cite{NOCS}. Several approaches \cite{CRNet}\cite{SGPA}\cite{SAR} leverage the shape prior and optimize the network design to achieve higher pose estimation accuracy. In those shape prior-based methods, two main tasks are included, which are deforming the shape prior to reconstruct the scale-invariant CAD model of target object in the canonical space (deformation task) and predicting the representation of point cloud of target object in the canonical space by taking the deformed shape prior as reference (registration task). In the training process, losses for the two tasks are summed together to optimize the network, which may lead to trade-off of network optimization and sub-optimal performance for each task. Furthermore, since the registration task takes the deformed shape prior as reference, there is inherent cascaded relation between two tasks. To this end, our DR-Pose is proposed in a two-stage design and training manner. In this two-stage pipeline, a completion-aided deformation network is first trained to achieve optimal deformation performance. Then a scaled registration network is subsequently trained with the inference output from the well trained deformation network.



As illustrated in \cref{teaser}, the first stage of our pipeline involves utilizing a completion network to recover the unseen parts of target object by taking the observed partial point cloud as input. Guided by the completed point cloud, the categorical shape prior is then deformed to reconstruct the scale-invariant CAD model of the target object.

In the second stage, taking the deformed shape prior from the well-trained deformation network model, a novel scaled registration network is designed to predict the correspondence matrix that best maps partial point cloud into canonical space. There are three main steps involved in the flow of this stage. Firstly, a shared KPFCN \cite{thomas2019KPConv} backbone is utilized to extract the features from partial point clouds and deformed shape prior. Secondly, these two sets of features are enhanced through a position-aware transformer block. Finally, the dot product is implemented on the two sets of enhanced features to output the correspondence matrix, which is further adjusted by the predicted scaling factors.

In summary, the main contributions of this work are:
\begin{itemize}
    \item We study the impact of a two-stage training approach in comparison to existing one-stage approaches. Based on this impact, a novel two-stage pipeline is proposed to separately optimize network design for specific deformation and registration tasks.
    \item We propose a novel completion-aided deformation network, where the missing part of target object is recovered and further guides the deformation task. With the help of point cloud completion, the deformation network achieves higher reconstruction accuracy.
    \item We propose a novel scaled registration network, in which pose-sensitive features are extracted for point cloud registration task. Concurrently scaling factors are predicted to further adjust the registration results and improve the pose estimation accuracy.
    \item Notably, extensive experimental results show our DR-Pose outperforms state-of-the-art shape prior-based approaches on the benchmark dataset of real-world REAL275 and synthetic CAMERA25, especially in stricter metrics.
\end{itemize}









\section{Related Work}
\subsection{Instance-level 6D Object Pose Estimation}
Instance-level pose estimation involves estimating the 6D pose of a specific 3D CAD object instance. There are generally three different approaches used in instance-level pose estimation methods. The first approach involves directly extracting embedding features to regress the 6D pose \cite{wang_densefusion_2019-1}\cite{xiang2018posecnn}. The second approach uses 2D-3D \cite{PVNET}\cite{pvn3d}\cite{CDPN}\cite{SO-Pose}\cite{pix2Pose} or 3D-3D \cite{pointposenet} correspondences  correspondences to solve a PnP \cite{pnp}\cite{epnp} problem and obtain the 6D pose. The third approach involves extracting a latent embedding to represent the object for retrieval purposes \cite{self6D}. However, all of these methods require precise 3D CAD instance models for both training and testing. This limitation makes these methods unsuitable for real-world scenarios where precise CAD models are unknown.

\subsection{Category-level 6D Object Pose Estimation}
The primary challenge in category-level pose estimation is intra-class variation, which makes it more difficult compared to instance-level pose estimation. Recent works have explored two main approaches to address this challenge. The first \cite{fs-net} approach involves using shape-based features to estimate two perpendicular vectors for direct 6D pose recovery. Following FS-Net, GPV-Pose \cite{GPV} further improves performance by introducing consistency loss between 3D bounding boxes, reconstruction, and pose. The second approach predicts the NOCS representation of the observed object point cloud and solve the 6D pose and size with Umeyama \cite{Umeyama} algorithm. SPD \cite{SPD} builds categorical shape prior within each category to provide reference for the prediction of NOCS representation. CR-Net \cite{CRNet} improves the deformation of shape prior by iterative refinement. SGPA \cite{SGPA} enriches the extracted features with transformer to further exchange information between target object and shape prior. However, those shape prior-based methods are still designed in single-stage format, which may lead to sub-optimal results in both tasks. In this paper, we propose a two-stage deformation-and-registration pipeline to optimize the network design and performance for each task.

\begin{figure*}[t]
	\centering
	\includegraphics[width=\linewidth]{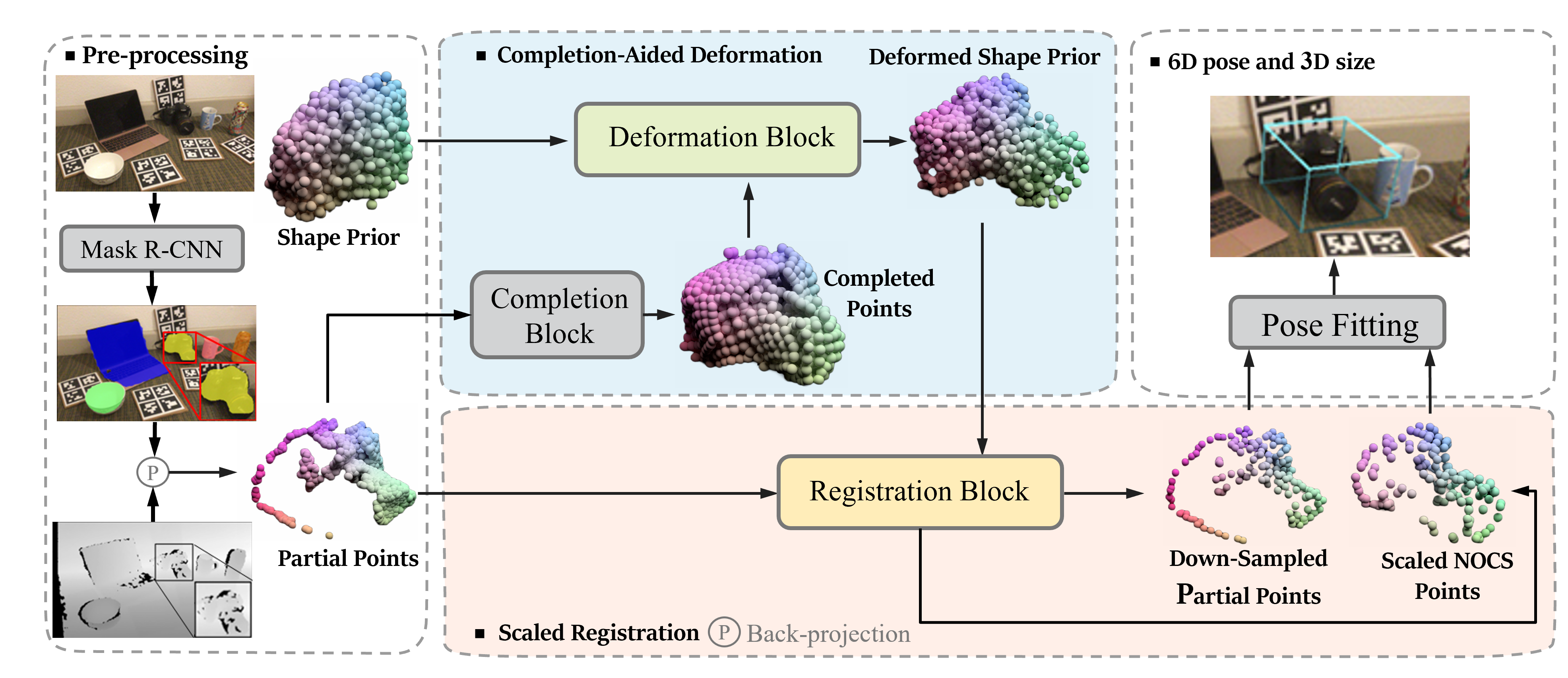}
	\caption{Overview of our two-stage deformation-and-registration pipeline. The Completion-aided deformation stage (\cref{deformation}) takes input of shape prior and observed partial point cloud to complete the missing part of object and guide the reconstruction for scale-invariant instance CAD model. The scaled registration stage (\cref{registration}) predicts representation of observed points in NOCS. The final 6D object pose and size are recovered through Umeyama algorithm \cite{Umeyama} between observed partial point cloud and its NOCS representation.}
	\label{pipeline}
\end{figure*}


\section{DR-Pose}
\subsection{Problem Formulation}
Given a calibrated RGB-D image, our goal is to estimate the 6D pose and the 3D size for an object of interest with respect to camera coordinate. The 6D object pose is represented as $\{R,T\}\in{SE(3)}$, where 3D rotation is $R \in{SO(3)}$ and 3D translation is $T\in{\mathbb{R}^{3}}$, and the 3D size is represented as $s \in{\mathbb{R}^{3}}$.
\subsection{Framework Overview}
The overview of our DR-Pose is depicted in \cref{pipeline}. First, an off-the-shelf segmentation algorithm (Mask-RCNN \cite{maskrcnn}) is used to segment out the objects of interest from input image. Then, a partial point cloud $O \in \mathbb{R}^{{N_{o}} \times 3}$ is generated for each target object by back-projecting the segmented depth image with camera intrinsic parameters.
In the completion-aided deformation stage (\cref{deformation}), a point cloud completion network is first utilized to generate the unseen part of $O$, denoted as $O_{com}$. Then the generated points $O_{com}$ and category shape prior $P \in{\mathbb{R}^{{N_{p}}\times 3}}$ are passed through a deformation network to reconstruct the scale-invariant CAD model denoted as deformed shape prior $P_{def}\in{\mathbb{R}^{{N_{p}}\times 3}}$. In the scaled registration stage (\cref{registration}), $O$ and $P_{def}$ are down-sampled to $\hat{O} \in \mathbb{R}^{{N_{\hat{o}}} \times 3}$ and $\hat{P}_{def} \in \mathbb{R}^{{N_{\hat{p}}} \times 3}$ respectively for efficient feature extraction and enhancement, and a correspondence matrix $A\in \mathbb{R}^{{N_{\hat{o}}} \times {N_{\hat{p}}}}$ is then estimated to match the two point clouds $\hat{O}$ and $\hat{P}_{def}$. The corresponding representation of $\hat{O}$ in canonical space, denoted as $O_{nocs}\in \mathbb{R}^{{N_{\hat{o}}} \times 3}$, is then computed by multiplying $A$ with deformed shape prior $\hat{P}_{def}$ and further adjusted with the predicted scaling factors. Finally, object 6D pose and size are computed by Umeyama \cite{Umeyama} algorithm between $\hat{O}$ and $O_{nocs}$.

\subsection{Completion-aided Deformation Stage} \label{deformation}
\begin{figure*}[t]
	\centering
	\includegraphics[width=\linewidth]{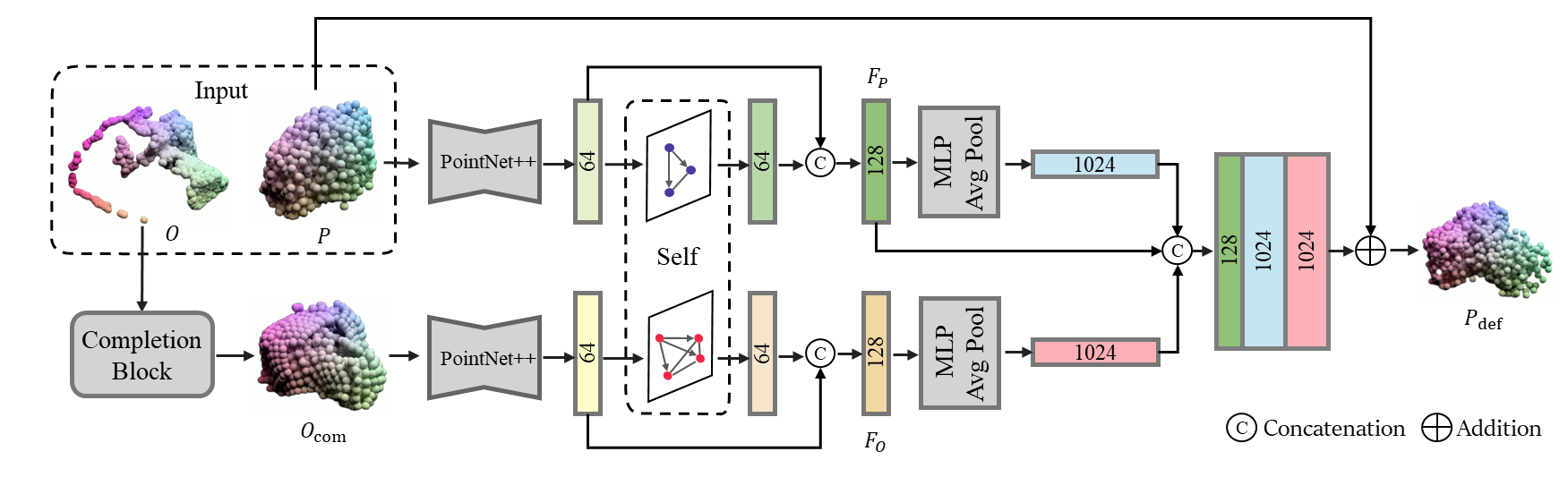}
	\caption{Overview of completion-aided deformation stage.}
	\label{deform_stage}
\end{figure*}

Taking the partial point cloud of target object as reference, the goal of this stage is to deform the categorical shape prior to reconstruct the scale-invariant CAD model of target object in the canonical space.  The architecture of completion-aided deformation stage is shown in \cref{deform_stage}.

\textbf{Point Cloud Completion.}  Given the segmented instance mask from RGB image, observed partial point cloud $O$ is back-projected from the masked pixels on the depth image with camera intrinsic parameters. However, the back-projected point cloud $O$ usually contains outliers due to the imperfect segmentation mask. Moreover, raw points directly captured by RGB-D cameras are usually sparse and incomplete owing to the limited sensor resolution and occlusions in depth image. To remove the outliers while retaining the overall shape and extract more pose-sensitive features from the object, an off-the-shelf point cloud completion network (PoinTr \cite{yu2021pointr}) is leveraged to complete the unseen part of the object.  With partial point cloud $O$ as input, PoinTr performs multi-scale point cloud generation, as it first predicts $N_{C}$ center points of the unseen parts. Then $N$ points are generated in the neighborhood of each center point.

\textbf{Scale-invariant CAD Model Reconstruction.} After the completion process, the generated point cloud $O_{com}$ provides more detailed object geometric features to facilitate the reconstruction of scale-invariant instance CAD model in canonical space. PointNet++ \cite{pointnet++} encoder is utilized to extract geometric features from the generated point cloud ${O_{com}}$ and category prior ${P}$ respectively, denoted as $F_{o}$ and $F_{p}$. The extracted features $F_{o}$ and $F_{p}$ are fed to a self-attention layer for feature enhancement. Average pooling is then used to extract global features $G_{o}$ and $G_{p}$ from the enhanced features. By concatenating $F_{p}$, $G_{o}$, and $G_{p}$, we aggregate local and global features embedding for each point of category prior $P$. Finally, per-point feature embeddings are passed through an MLP to regress per-point deformation field $D \in \mathbb{R}^{{N_{p}} \times 3}$. The scale-invariant CAD model $P_{def}$ for the target object is reconstructed by adding this deformation field $D$ with initial shape prior $P$:
\begin{equation}
    P_{def} = P + D
\end{equation}
\subsection{Scaled Registration Stage} \label{registration}
\begin{figure*}[t]
	\centering
	\includegraphics[width=\linewidth]{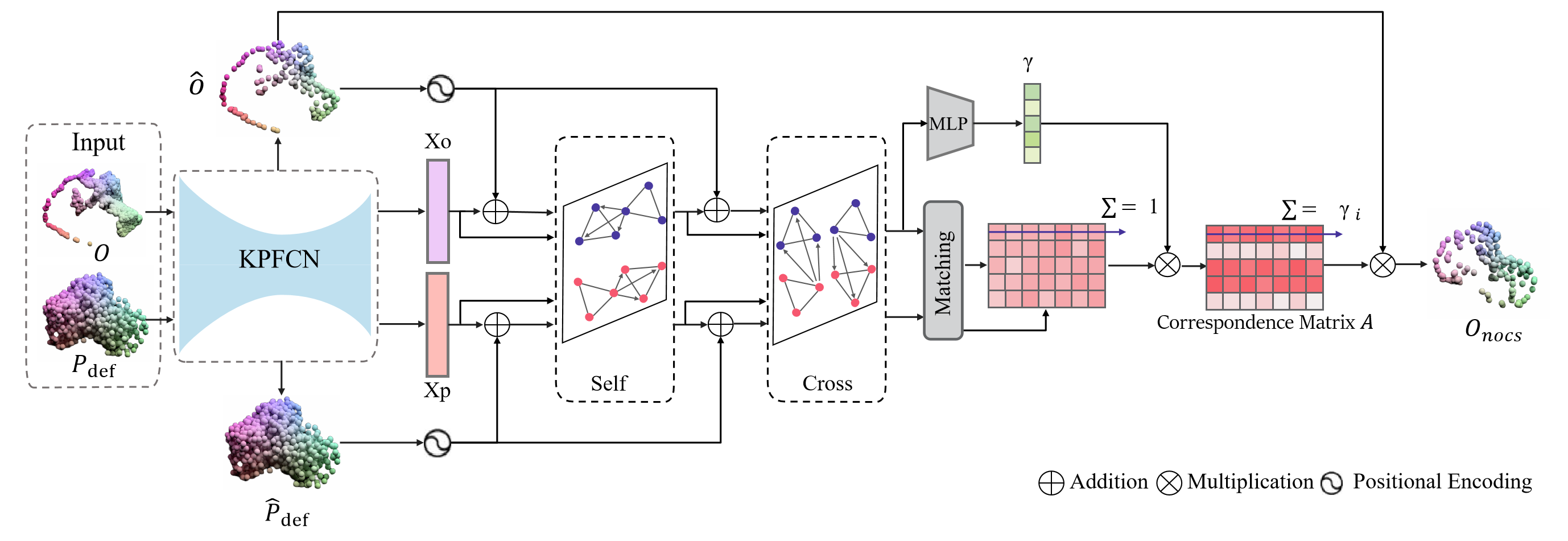}
	\caption{Overview of scaled registration stage.}
	\label{registration_stage}
\end{figure*}

With deformed prior $P_{def}$ from the first stage, the goal of this stage is to take $P_{def}$ as reference and predict the coordinates of each point of the observed point cloud $O$ in the canonical space.
The architecture of scaled registration stage is shown in \cref{registration_stage}.

\textbf{Local Geometric Feature Extraction.}
Given the deformed prior $P_{def}$ and observed partial point cloud $O$, KPFCN \cite{thomas2019KPConv} is utilized as a feature encoder to extract geometric features in this stage. For efficient computation in the subsequent process (feature enhancement and matching), we remove the decoder blocks from the 2nd up-sampling layer to the end. In this way, input points are down-sampled to $\hat{O}$ and $\hat{P}_{def} \in \mathbb{R}^{{N_{\hat{p}}} \times 3}$ respectively. Then local geometric features extracted from the down-sampled points are denoted as $x_{o}\in \mathbb{R}^{{N_{\hat{o}}} \times d}$ and $x_{p}\in \mathbb{R}^{{N_{\hat{p}}} \times d}$ for $\hat{O}$ and $\hat{P}_{def}$ respectively, in which $d$ represents feature dimension.

\textbf{Positional Encoding.} The extracted feature from the KPFCN backbone may lack spatial information, which can cause confusion when dealing with symmetric or repeated geometric features. To address this issue, we utilize Sinusoidal positional encoding (PE) to embed positional information into the extracted feature to clarify the ambiguity:

\begin{equation}
    x_{o}^{i} \gets \Theta(\hat{O}_{i})+x_{o}^{i} \ \ \ \ x_{p}^{j} \gets \Theta(\hat{P}_{def}^{j})+x_{p}^{j}
\end{equation}

where $\Theta$ represents positional encoding, $\hat{O}_{i}$ and $\hat{P}_{def}^{j}$ represent each point in $\hat{O}$ and $\hat{P}_{def}$ respectively.

\textbf{Transformer-based Feature Enhancement.} With local features $x_{o}$ and $x_{p}$ extracted by KPFCN and encoded with positional encoding, we then leverage a transformer block to aggregate contextual cues intuitively within each group of point clouds by self-attention operation and exchange information between two groups of point clouds for subsequent correspondence prediction with the cross-attention operation. The attention operation measures the similarity between query vector \textbf{q} and key vector \textbf{k}, and the output is the weighted sum of the value vector \textbf{v} based on the similarity scores. 


\begin{equation}
    q_{i}=W_{q}{x_{o}^{i}}\ \ \ k_{j}=W_{k}{x_{o}^{j}} \ \ \ v_{j}=W_{v}{x_{o}^{j}}
\end{equation}
where $W_{q}$, $W_{k}$, $W_{v} \in{\mathbb{R}^{d\times d}}$ are learnable projection matrices. The feature $x_{o}^{i}$ is finally updated by

\begin{equation}
    x_{o}^{i}\gets x_{o}^{i}+MLP(cat[q_{i},\sum_{j}{w_{ij}v_{j}}])
\end{equation}
Where $w_{ij} = softmax(q_{i}k_{j}^{T}/\sqrt{d})$ is normalized similarity score, $MLP(\cdot)$ denotes a shared multi-layer-perceptron (MLP), and $cat[\cdot,\cdot]$ is the concatenation operator.


 \textbf{Scaled Registration.} \label{scale_explain} With the enhanced feature $x_{o}$ and $x_{p}$, we perform element-wise dot product between two point clouds and obtain a similarity scoring matrix $S\in\mathbb{R}^{{N_{\hat{O}}}\times{N_{\hat{P}}}}$:

\begin{equation}
    S(i,j) = \frac{1}{\sqrt{d}}	\left \langle {W_{o}}{x_{o}^{i}}, {W_{p}}{x_{p}^{j}}	\right \rangle
\end{equation}
where $W_{o}, W_{p} \in{\mathbb{R}^{{d} \times {d}}}$ are learnable projection matrices. Softmax is then applied to each row of the scoring matrix $S_{i}$ to obtain the correspondence matrix $A \in{\mathbb{R}^{{N_{\hat{o}}} \times {N_{\hat{p}}}}}$. Since each row of A sums to 1, it represents a soft correspondence between observed partial point cloud $\hat{O}$ and deformed prior $\hat{P}_{def}$.
Finally we apply the correspondence matrix $A$ on the deformed shape prior $\hat{P}_{def}$ to obtain the representation of observed partial point cloud $\hat{O}$ in canonical space as {$O_{nocs}\in{\mathbb{R}^{N_{\hat{O}\times{3}}}}$}:
\begin{equation}
    O_{nocs}^{i}= \left \langle A_{i}, \hat{P}_{def} \right \rangle =\sum\limits_{j=1}^{N_{\hat{p}}}{a_{ij} \hat{P}_{def}^{j}}
\end{equation}
where $\left \langle \cdot, \cdot \right \rangle$ denotes dot product, $O_{nocs}^{i}$ represents the corresponding point in canonical space of each observed point, $A_{i}$ represents each row of the correspondence matrix, and $\hat{P}_{def}^{j}$ represents each down-sampled deformed prior point.

Nevertheless, the deformation process from $P$ to $P_{def}$ can not be imperfect. Since each row of the correspondence matrix $A$ sums to 1, if there is error w.r.t translation, rotation or scale between the deformed shape prior and the ground truth CAD model, then the weighted sum of coordinates of the deformed prior $P_{def}$ will never reach the ground truth coordinates for some parts of the object no matter the values of correspondence matrix $A$.
In order to alleviate this limitation, we add a branch at the head of registration network to regress a vector of scaling factors $\gamma\in{\mathbb{R}^{N_{\hat{o}\times1}}}$. By applying scaling factors to each row of correspondence matrix $A$ as:
\begin{equation}
    a_{ij} \gets \gamma_{i} a_{ij}
\end{equation}
the sum of each row becomes $\sum\limits_{j=1}^{N_{\hat{p}}} a_{ij} = \gamma_{i}$ instead of 1. These learnable scaling factors enable our DP-Pose to adjust the predicted $O_{nocs}$ to break the limit of imperfect $P_{def}$.

\textbf{Pose Fitting.}
Given the observed partial point cloud, $\hat{O}$ and its corresponding $O_{nocs}$, the optimal similarity transformation parameters (rotation, translation, and scaling) can be computed by solving the absolute orientation problem using Umeyama algorithm \cite{Umeyama}.
\subsection{Loss Function}

\textbf{Completion Loss.} We adopt Chamfer Distance (CD) with L1-norm to supervise the completion quality of the PoinTr network, which is introduced in \cite{yu2021pointr}.

\begin{table*}[t]
	\captionsetup{justification=centering}
	\renewcommand\arraystretch{1.0}
	\begin{center}
	\caption{Comparison with state-of-the-art methods on CAMERA25 and REAL275 datasets. Overall best results are in bold.}
		\label{Camera25_Real275}
		\begin{tabular}{p{1.9cm}<{\centering}|p{0.8cm}<{\centering}|p{0.8cm}<{\centering}|p{0.8cm}<{\centering}|p{0.8cm}<{\centering}|p{0.9cm}<{\centering}|p{0.9cm}<{\centering}|p{0.8cm}<{\centering}|p{0.8cm}<{\centering}|p{0.8cm}<{\centering}|p{0.8cm}<{\centering}|p{0.9cm}<{\centering}|p{0.9cm}<{\centering}}
		\hline
		\hline
		& \multicolumn{6}{c|}{CAMERA25} & \multicolumn{6}{c}{REAL275} \\
        \hline
        Method & $3D_{50}$ & $3D_{75}$ & $5^\circ2cm$ & $5^\circ5cm$ & $10^\circ2cm$& $10^\circ5cm$ & $3D_{50}$ & $3D_{75}$ & $5^\circ2cm$ & $5^\circ5cm$& $10^\circ2cm$& $10^\circ5cm$\\
        \hline
        NOCS\cite{NOCS} & 83.9 & 69.5 & 32.3 & 40.9 & 48.2 & 64.6 & 78.0 & 30.1 & 7.2 & 10.0 & 13.8 & 25.2\\
        SPD\cite{SPD} &  93.2 & 83.1 & 54.3 & 59.0 & 73.3 & 81.5 & 77.3 & 53.2 & 19.3 & 21.4 & 43.2 & 54.1 \\
        SGPA\cite{SGPA} & 93.2 & 88.1 & 70.7 & 74.5 & 82.7 & 88.4 & 80.1 & 61.9 & 35.9 & 39.6 & 61.3 & 70.7\\
        CR-Net\cite{CRNet} & \textbf{93.8} & 88.0 & 72.0 & 76.4 & 81.0 & 87.7 & \textbf{79.3} & 55.9 & 27.8 & 34.3 & 47.2 & 60.8\\
        SAR-Net\cite{SAR} & 86.8 & 79.0 & 66.7 & 70.9 & 75.3 & 80.3 & \textbf{79.3} & 62.4 & 31.6 & 42.3 & 50.3 & 68.3\\
        Ours & 92.7 & \textbf{89.3} & \textbf{74.3} & \textbf{78.0} & \textbf{85.0} & \textbf{89.7} & 78.9 & \textbf{68.2} & \textbf{41.7} & \textbf{46.0} & \textbf{67.7} & \textbf{76.3}\\
			\hline
		\end{tabular}
	\end{center}
\end{table*}





\textbf{Deformation Loss.} Following SPD \cite{SPD}, we indirectly supervise the deformation field D by measuring the accuracy of scale-invariant CAD model reconstruction with L2-normed Chamfer Distance:
\begin{equation}
        \mathcal{L}_{cd} = \sum_{x\in{P_{def}}}^{}\mathop{min}\limits_{y\in{P_{gt}}}{\left\lVert x - y \right\rVert}_{2}^{2} + \sum_{y\in{P_{gt}}}^{}\mathop{min}\limits_{x\in{P_{def}}}{\left\lVert y - x \right\rVert}_{2}^{2}
\end{equation}
where $P_{gt}$ is the scale-invariant instance CAD model given in the training set.

Additionally, a regularization loss to discourage large deformation in the deformation field D:

\begin{equation}
    \mathcal{L}_{delta}=\frac{1}{N_{p}}\sum_{d_{i}\in{D}}{\left\lVert d_{i} \right\rVert}_{2}
\end{equation}

The overall loss for the first stage is: 
\begin{equation}
    \mathcal{L}_{def} = \lambda_{0}\mathcal{L}_{cd} + \lambda_{1}\mathcal{L}_{delta}
\end{equation}

\textbf{Registration Loss.} We apply the correspondence matrix $A$ on deformed shape prior $\hat{P}_{def}$ to obtain a NOCS coordinate prediction $O_{nocs}$ for each observed point cloud $\hat{O}$. Since the corresponding ground-truth NOCS coordinate $O_{nocs}^{gt}$ is given in the training set, we supervise the correspondence matrix A by constraining the distance between the predicted NOCS coordinates $\textbf{x}$ and the ground-truth ones $\textbf{y}$. The correspondence loss $\mathcal{L}_{corr}$ is defined as:


\begin{equation}
    \mathcal{L}_{corr}(\textbf{x},\textbf{y})=\frac{1}{N_{\hat{o}}}\left \{ \begin{array}{cc}
        5(\textbf{x}-\textbf{y})^{2} & \left| \textbf{x} - \textbf{y} \right| \leq{0.1} \\
         \left| \textbf{x} - \textbf{y} \right|-0.05 & otherwise
    \end{array}
    \right.
\end{equation}
where $\textbf{x} = (x_{1},x_{2},x_{3})\in{O_{nocs}}$, and $\textbf{y} = (y_{1},y_{2},y_{3}) \in {O_{nocs}^{gt}}$.

During the training process, we both calculate $\mathcal{L}_{corr}$ loss before and after applying scaling factors to adjust NOCS coordinates $O_{nocs}$, denoted as $\mathcal{L}_{corr0}$ and $\mathcal{L}_{corr1}$ respectively. The $\mathcal{L}_{corr}$ loss is the weighted sum of $\mathcal{L}_{corr0}$ and $\mathcal{L}_{corr1}$:

\begin{equation}
    \mathcal{L}_{corr} = \lambda_{2}  \mathcal{L}_{corr0} + \lambda_{3} \mathcal{L}_{corr1}
\end{equation}

In addition, the same regularization loss $L_{entropy}$ in SPD \cite{SPD} is adopted to encourage each row of A to be a peaked distribution.
Finally, the loss for this stage is the weighted sum of two losses: 
\begin{equation}
    \mathcal{L}_{regis} = \lambda_{4}\mathcal{L}_{corr} + \lambda_{5}\mathcal{L}_{entropy}
\end{equation}

\begin{figure*}[t]
	\centering
	\includegraphics[width=0.9\linewidth]{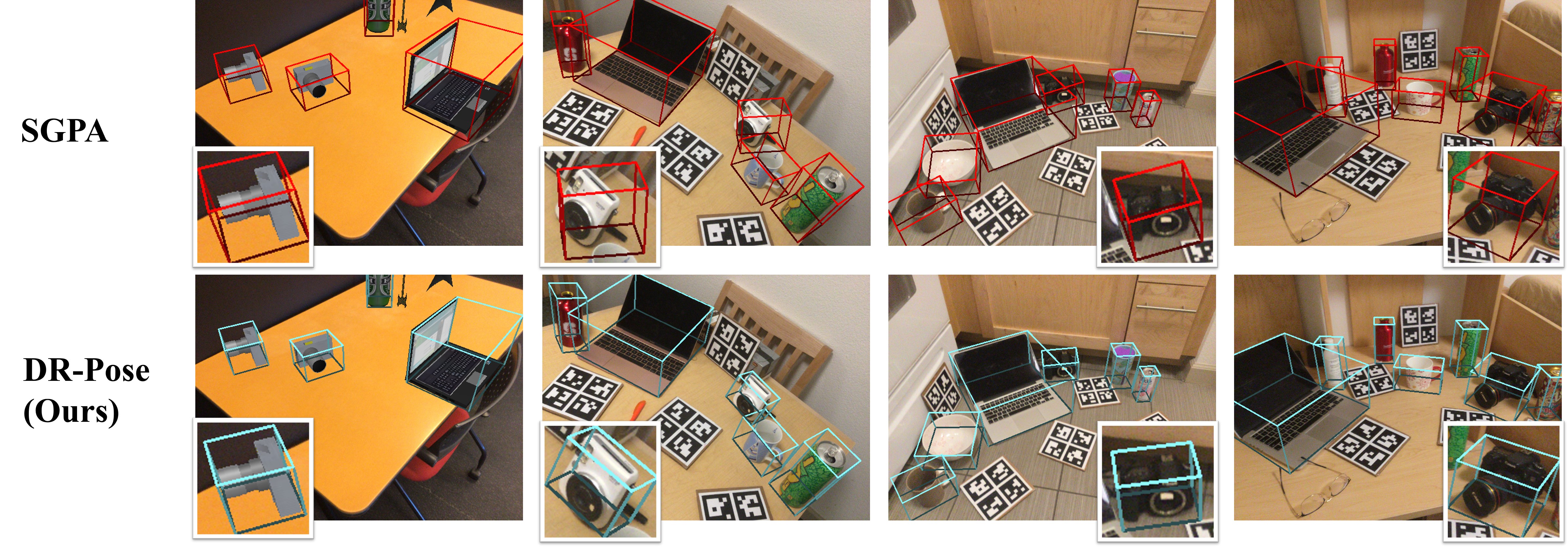}
	\caption{Qualitative comparisons between our DR-Pose and SGPA \cite{SGPA} on CAMERA25 and REAL275 datasets. The estimated 6D Pose and size is visualized as the tight-oriented bounding box around the target instance.}
	\label{scene}	
\end{figure*}

\section{Experiments}
\subsection{Dataset}
We use NOCS \cite{NOCS} benchmarks to train and evaluate our DR-Pose. It consists of two datasets: the synthetic dataset CAMERA25 and the real-world dataset REAL275. CAMERA25 contains 300K RGB-D images with rendered objects and virtual backgrounds, in which 25K images are set aside for testing. REAL275 contains 8K RGB-D images, in which 2.75K images are used for testing. Two datasets share the same 6 categories including \textit{bottle}, \textit{bowl}, \textit{camera}, \textit{can}, \textit{laptop}, and \textit{mug}.

\subsection{Evaluation Metrics}

Following SPD \cite{SPD}, we independently compute mean Average Precision (mAP) of 3D Intersection-Over-Union (IoU) for 3D object detection and $\boldsymbol{m}^{\circ}\boldsymbol{n}cm$ for 6D object pose estimation.

\textbf{3D IoU.} It measures the overlap between the predicted object 3D bounding box and the ground truth bounding box at thresholds of 50\%, 75\%.

\textbf{$\boldsymbol{m}^{\circ}\boldsymbol{n}cm$.} A more straightforward metric that directly compares errors in translation and rotation between the predicted pose and the ground truth pose. A estimated pose is thereby considered correct if the translation and rotation errors are both below the given thresholds. We adopt $\boldsymbol{5}^{\circ}\boldsymbol{2}cm$, $\boldsymbol{5}^{\circ}\boldsymbol{5}cm$, $\boldsymbol{10}^{\circ}\boldsymbol{2}cm$, $\boldsymbol{10}^{\circ}\boldsymbol{5}cm$ metrics to evaluate the pose estimation accuracy.

\subsection{Implementation Details} Our DR-Pose is trained on CAMERA25 and REAL275 in a mixed way following \cite{SGPA}. 
For a fair comparison with \cite{SGPA}, we use the same instance segmentation results from Mask-RCNN to back-project the instance point cloud. Then PoinTr network is trained to implement point cloud completion for the randomly down-sampled observed partial point cloud with 1024 points. The generated point cloud contains 1152 points. For synthetic data with few noise, the initial 1024 points are concatenated behind generated points. While for real-world noisy data, only the generated points are passed to the subsequent task. Category shape prior is obtained from the auto-encoder network in \cite{SPD}, consisting of 1024 points each prior. In the training of the registration network, random down-sampling is not implemented on $O$ as in the deformation stage since radius down-sampling is implemented in KPFCN for input points.
The parameters for all loss terms $\{\lambda_{0}, \lambda_{1}, \lambda_{2}, \lambda_{3}, \lambda_{4}, \lambda_{5}\} = \{5.0, 0.01, 0.6, 0.4, 1.0, 0.0001\}$.
DR-Pose is trained on a single NVIDIA RTX3090 GPU. The deformation network is trained with a batch size of 16 for 100 epochs. The registration network is trained with a batch size of 8 for 150 epochs.

\subsection{Comparison with State-of-the-Art Methods}
We compare our DR-Pose with NOCS \cite{NOCS} and shape prior-based state-of-the-art methods SPD \cite{SPD}, SGPA \cite{SGPA},
CR-Net \cite{CRNet} and SAR-Net \cite{SAR} on CAMERA25 and REAL275 datasets. Quantitative results in \cref{Camera25_Real275} shows the superiority of our proposed DR-Pose on both datasets, especially for the strictest metrics such as $3D_{75}$ and $5^\circ2cm$.

\textbf{CAMERA25.} On this benchmark dataset, DR-Pose achieves state-of-the-art results on 5 out of 6 metrics and outperforms our baseline SGPA \cite{SGPA}  by 1.2\% and 2.3\% respectively in terms of the strictest metrics $3D_{75}$ and $5^\circ2cm$. Though DR-Pose fails to achieve best result on the less strict metric  $3D_{50}$, there is only marginal gap compared to the state-of-the-art result.

\textbf{REAL275.} The improvement on REAL275 is particularly significant compared to the improvement on CAMERA25, indicating the effectiveness of the proposed pipeline in real-world scenarios. On this benchmark dataset, DR-Pose also achieves state-of-the-art results on 5 out of 6 metrics. Specifically, for the two strictest metrics $3D_{75}$ and $5^\circ2cm$, DR-Pose improves the state-of-the-art results by a large margin as it outperforms the baseline method SGPA by 6.3\% and 5.8\% respectively.

Qualitative comparison with SGPA \cite{SGPA} on the CAMERA25 and REAL275 datasets is shown in \cref{scene}. 

\subsection{Ablation Studies} 
We conducted four ablation studies to justify our choice of design for the DR-Pose pipeline, evaluating the pipeline's performance on the REAL275 datasets. As the 3D IoU metric can be influenced by multiple factors, including translation error, rotation error, rotation axis, and estimated size, there may not always be a positive correlation between 3D IoU and $\boldsymbol{m}^{\circ}\boldsymbol{n}cm$. In fact, most of the previous works \cite{NOCS}\cite{SPD}\cite{CRNet}\cite{SGPA}\cite{SAR} have reported discrepancies between trends of 3D IoU and $\boldsymbol{m}^{\circ}\boldsymbol{n}cm$ metrics in their ablation studies. In such cases, the strictest $\boldsymbol{m}^{\circ}\boldsymbol{n}cm$ metric reported in the paper is typically used as a reference to decide the network design, to ensure high accuracy in pose estimation task. Therefore, following this convention, we used $\boldsymbol{5}^{\circ}\boldsymbol{2}cm$ as a reference to determine the network design for DR-Pose in the ablation studies.


\begin{table}[t]
        \small
	\captionsetup{justification=centering}
	\renewcommand\arraystretch{1.0}
	\begin{center}
		\caption{Comparison of performance between single-stage and reproduced two-stage SGPA network}
		\label{compare_two_stage}
		\begin{tabular}{p{1.2cm}<{\centering}|p{0.6cm}<{\centering}|p{0.5cm}<{\centering}|p{0.8cm}<{\centering}|p{0.4cm}<{\centering}|p{0.7cm}<{\centering}|p{0.5cm}<{\centering}|p{0.5cm}<{\centering}}
        \hline
        \hline
        \multirow{2}*{Method} & \multicolumn{7}{c}{Chamfer distance error ($\times10^{-3}$) on REAL275 } \\
        \cline{2-8}
        & bottle & bowl & camera & can & laptop & mug & mean \\
        \hline
        SGPA & 2.93 & \textbf{0.89} & \textbf{5.51} & 1.75 & 1.62 & \textbf{1.12} & 2.44 \\
        SGPA* & \textbf{2.87} & 0.97 & 6.16 & \textbf{1.47} & \textbf{1.24} & 1.21 & \textbf{2.32} \\ 
        \hline
		\end{tabular}
  \begin{tabular}{p{1.2cm}<{\centering}|p{1.33cm}<{\centering}|p{1.33cm}<{\centering}|p{1.33cm}<{\centering}|p{1.33cm}<{\centering}}

        \multirow{2}*{Method} & \multicolumn{4}{c}{Pose estimation accuracy (\%) on REAL275} \\
        \cline{2-5}
        & $5^\circ2cm$ & $5^\circ5cm$ & $10^\circ2cm$& $10^\circ5cm$ \\
        \hline
        SGPA & 35.9 & 39.6 & 61.3 & 70.7 \\
        SGPA* & \textbf{37.8} & \textbf{42.1} & \textbf{66.9} & \textbf{76.0} \\
        \hline
		\end{tabular}
	\end{center}
\end{table}
\begin{table}[t]
        \small
	\captionsetup{justification=centering}
	\renewcommand\arraystretch{1.0}
	\begin{center}
		\caption{Comparison of the model reconstruction accuracy in CD metric ($\times10^{-3}$).}
		\label{compare_CD_metric}
		\begin{tabular}{p{1.2cm}<{\centering}|p{0.6cm}<{\centering}|p{0.5cm}<{\centering}|p{0.8cm}<{\centering}|p{0.4cm}<{\centering}|p{0.7cm}<{\centering}|p{0.5cm}<{\centering}|p{0.5cm}<{\centering}}
        \hline
        \hline
        \multirow{2}*{Method} & \multicolumn{7}{c}{Chamfer distance error ($\times10^{-3}$) on REAL275} \\
        \cline{2-8}
        & bottle & bowl & camera & can & laptop & mug & mean \\
        \hline
        SPD & 3.44 & 1.21 & 8.89 & 1.56 & 2.91 & \textbf{1.02} & 3.17 \\
        CR-Net & 2.99 & 0.96 & 7.57 & \textbf{1.31} & 1.25 & 1.19 & 2.53 \\
        CR-Net* & 2.10 & 1.04 & 6.58 & 3.07 & 1.45 & 1.39 & 2.60 \\
        SGPA & 2.93 & \textbf{0.89} & 5.51 & 1.75 & 1.62 & 1.12 & 2.44 \\
        SGPA* & 2.87 & 0.97 & 6.16 & 1.47 & 1.24 & 1.21 & 2.32 \\ 
        $Ours\_w/o$ & 2.92 & 1.13 & 6.72 & 1.46 & 1.28 & 1.27 & 2.46 \\
        $Ours$ & \textbf{2.16} & 0.92 & \textbf{5.26} & 1.69 & \textbf{1.12} & 1.23 & \textbf{2.06} \\
        \hline
		\end{tabular}
	\end{center}
\end{table}
\begin{figure}[t]
	\centering
	\includegraphics[width=0.9\linewidth]{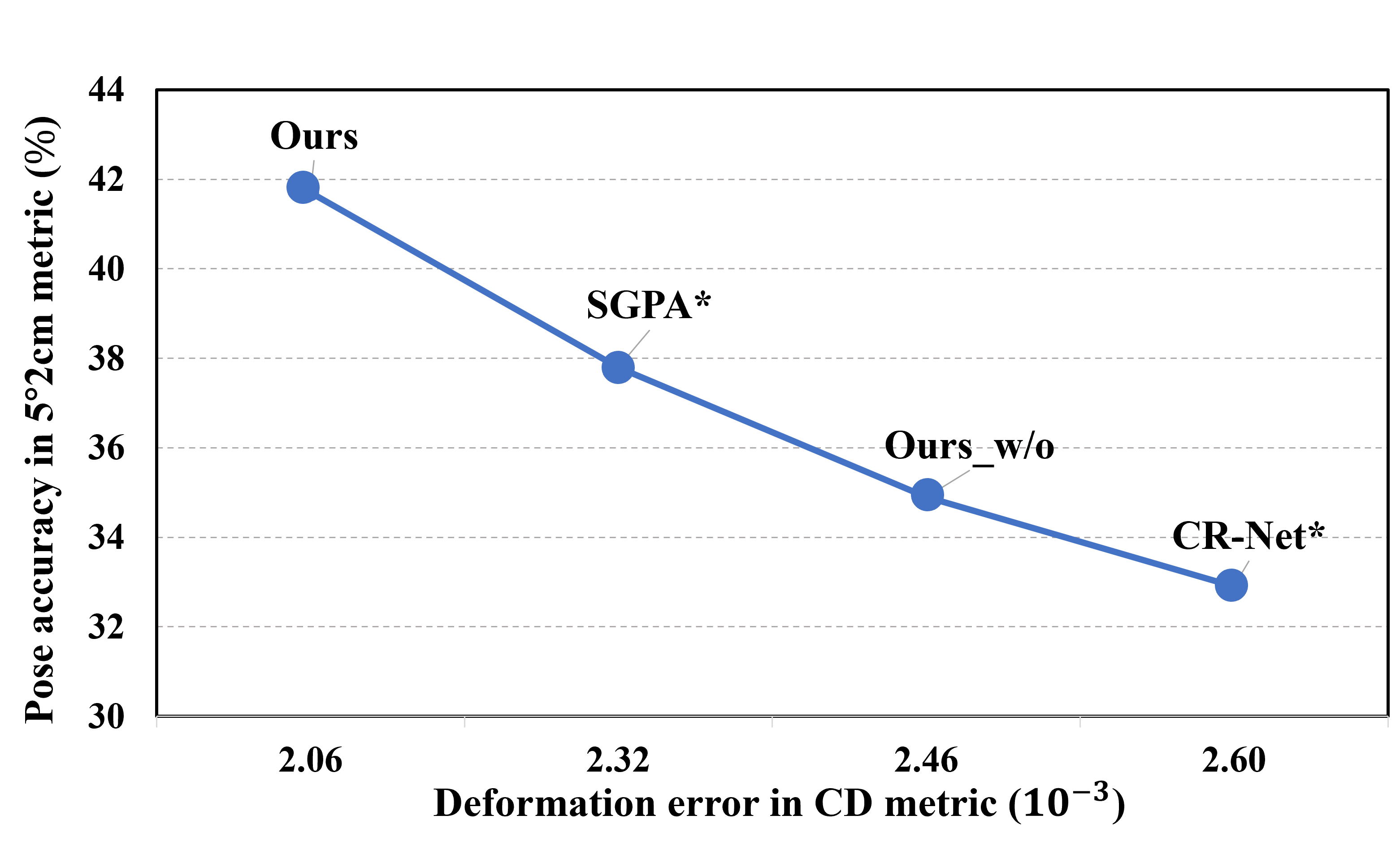}
	\caption{Visualization of CD error and pose estimation accuracy.}
	\label{trend}	
\end{figure}

\textbf{Advantage of Two-stage Pipeline over Single-stage.} In shape prior-based methods, deforming the shape prior and finding correspondence between two point clouds are distinct tasks supervised by different types of losses, which may lead to different learned features of network. Therefore a single-stage design and training format may lead to sub-optimal results in both tasks. To this end, we decouple the network design of SGPA \cite{SGPA} into a cascaded two-stage manner and train the deformation stage first. With the well-trained first stage model, we then train the registration network of SGPA. The comparison of result of each stage with the single-stage design reported in \cite{SGPA} are shown in \cref{compare_two_stage}. In the table, we denote the result of two-stage SGPA reproduced by us as SGPA*.
To evaluate the accuracy of scale-invariant CAD model reconstruction for target object, we report Chamfer Distance (CD) error between the deformed shape prior $P_{def}$ and the ground truth scale-invariant instance model. It is obvious that the performance of two-stage SGPA outperforms the original results in each stage.


\begin{table}[t]
        \small
	\captionsetup{justification=centering}
	\renewcommand\arraystretch{1.0}
	\begin{center}
		\caption{Pose estimation accuracy from the second stage after using different configurations in the first stage.}
		\label{stage2_results}
		\begin{tabular}{p{1.2cm}<{\centering}|p{1.32cm}<{\centering}|p{1.32cm}<{\centering}|p{1.32cm}<{\centering}|p{1.32cm}<{\centering}}
        \hline
        \hline
        \multirow{2}*{Method} & \multicolumn{4}{c}{Pose estimation accuracy (\%) on REAL275} \\
        \cline{2-5}
        & $5^\circ2cm$ & $5^\circ5cm$ & $10^\circ2cm$& $10^\circ5cm$ \\
        \hline
        CR-Net*  & 32.9 & 37.2 & 64.4 & 73.6 \\
        SGPA* & 37.8 & 42.1 & 66.9 & 76.0 \\
        $Ours\_{w/o}$ & 34.9 & 40.5 & 66.2 & 76.7\\
        $Ours_{4cat}$ & \textbf{41.7} & \textbf{46.0} & \textbf{67.7} & \textbf{76.3}\\
        \hline
		\end{tabular}
	\end{center}
\end{table}

    
\begin{table}[h!]
        \small
	\captionsetup{justification=centering}
	\renewcommand\arraystretch{1.0}
	\begin{center}
		\caption{Evaluation of how applying scaling factors affects pose estimation accuracy in the second stage.}
		\label{scale_factor}
		\begin{tabular}{p{1.2cm}<{\centering}|p{1.32cm}<{\centering}|p{1.32cm}<{\centering}|p{1.32cm}<{\centering}|p{1.32cm}<{\centering}}
        \hline
        \hline
        Scaling & \multicolumn{4}{c}{Pose estimation accuracy (\%) on REAL275} \\
        \cline{2-5}
        factors & $5^\circ2cm$ & $5^\circ5cm$ & $10^\circ2cm$& $10^\circ5cm$ \\
        \hline
        Not  & 37.0 & 41.1 & 66.6 & 75.2\\
        Apply  & \textbf{41.7} & \textbf{46.0} & \textbf{67.7} & \textbf{76.3} \\
        \hline
		\end{tabular}
	\end{center}
\end{table}

\textbf{Effect of Point Cloud Completion in Deformation Stage.}
To assess the impact of point cloud completion on the reconstruction of scale-invariant CAD models in the deformation stage, we report the Chamfer Distance (CD) error between the deformed prior $P_{def}$ and the ground truth scale-invariant instance model in \cref{compare_CD_metric}. In the table, SGPA* denotes the two-stage SGPA network reproduced by us and CR-Net* denotes the pre-trained model released by the authors reproduced on our side. We also denote the result of our deformation network without using completion as $Ours\_w/o$, while result of our completion-aided deformation network as $Ours$. Our experimental results indicate that implementing completion on only four categories (bottle, bowl, camera, and can) instead of all six categories can lead to the best result in the first stage. We observed that the handle of a mug is not always visible, which can confuse the completion network and generate an ambiguous completed point cloud. For laptops, since the object size is relatively large and most parts of the observed instance are visible in the dataset, applying completion can add noisy points and lower the performance of the deformation network.


\textbf{More Accurate Deformation, More Accurate Registration?} To support our claim that a more accurate deformation stage can lead to improved performance in the registration stage, we conduct an experiment where we keep the network design of the registration stage (second stage) fixed and train it using four different configurations of the deformation stage. These configurations include SGPA*, CR-Net*, $Ours\_w/o$, $Ours$, the deformation accuracy of each configuration is compared in \cref{compare_CD_metric}. The accuracy of second stage achieved under each configuration is compared in \cref{stage2_results}. In addition to the quantitative results, we also visualize the CD error and pose estimation accuracy for each configuration in \cref{trend}. It is evident from the visualization that the configuration with a more accurate deformation result also has higher pose accuracy. This trend demonstrates that improving the performance of first stage can also improve the overall performance of our pipeline, which aligns with our purpose of designing a two-stage pipeline.

\textbf{Effect of Scaling Factors.} As explained in \cref{scale_explain}, the registration network maps the observed object points into canonical space in a voting manner, taking the deformed shape prior from the first stage as reference. This is susceptible to the quality of deformation result. To further adjust the mapped result of each point, we utilize the enriched features extracted in the second stage to predict point-wise adjustment for the NOCS points. To evaluate performance without scaling factors, we compute correspondence loss as $L_{corr} = L_{corr0}$ in the training process. The comparison of results of using scaling factors or not is shown in \cref{scale_factor}. It is evident that applying scaling factors can further improve the pose estimation accuracy in our pipeline.




\section{Conclusion}
In this paper, we propose a two-stage deformation-and-registration pipeline for category-level 6D object pose estimation (DR-Pose). It consists of a completion-aided deformation stage and a scaled registration stage. In the first stage, unseen parts of target object are recovered with point cloud completion network, which then guides the deformation of shape prior in canonical space. In the second stage, a novel scaled registration network is designed to extract and enrich pose-sensitive features and scaling factors are predicted to further adjust the predicted NOCS representation. Extensive experiments on two well-acknowledged benchmarks show that our DR-Pose dramatically outperforms state-of-the-art shape prior-based competitors. Based on the two-stage design of DR-Pose, future works such as incorporating multiview inputs for occlusion scenes or refining network design for each task in the two stages can reduce the accumulated error and achieve higher overall performance.

\section*{ACKNOWLEDGMENT}

This research is supported by the Agency for Science, Technology and Research (A*STAR) under its AME Programmatic Funding Scheme (Project \#A18A2b0046).

{\bibliographystyle{IEEEtran}
\bibliography{root}
}
\addtolength{\textheight}{-12cm}   

\end{document}